\documentclass[11pt]{article}

\usepackage[final]{acl}

\usepackage{times}
\usepackage{latexsym}

\usepackage[T1]{fontenc}
\usepackage[utf8]{inputenc}

\usepackage{microtype}

\usepackage{inconsolata}

\usepackage{graphicx}
\usepackage{comment}

 \usepackage{float}

\usepackage{array}     % For defining custom column widths

\usepackage{multirow}
\usepackage{booktabs} % For thicker horizontal lines

\usepackage{makecell}
\PassOptionsToPackage{table,xcdraw}{xcolor}
\usepackage{xcolor}
\usepackage{colortbl}

\usepackage{amsmath} % for the \eqref command
\title{Leveraging Small LLMs for Argument Mining in Education: Argument Component Identification, Classification, and Assessment}

\author{
    \textbf{Lucile Favero\textsuperscript{1}},
    \textbf{Juan Antonio Pérez-Ortiz\textsuperscript{2}},
    \textbf{Tanja Käser\textsuperscript{3}},
   \textbf{ Nuria Oliver\textsuperscript{1}}
\\
    \textsuperscript{1}ELLIS Alicante, Spain,
    \textsuperscript{2}Universitat d'Alacant, Spain\,
    \textsuperscript{3}École Polytechnique Fédérale de Lausanne, EPFL, Switzerland
\\
 \small{
   \textbf{Correspondence:} \href{mailto:email@domain}{lucile@ellisalicante.org}
 }}

\begin{document}
\maketitle
\begin{abstract}
 Argument mining algorithms analyze the argumentative structure of essays, making them a valuable tool for enhancing education by providing targeted feedback on the students' argumentation skills. While current methods often use encoder or encoder-decoder deep learning architectures, decoder-only models remain largely unexplored, offering a promising research direction. 
 This paper proposes leveraging open-source, small Large Language Models (LLMs) for argument mining through few-shot prompting and fine-tuning. These models' small size and open-source nature ensure accessibility, privacy, and computational efficiency, enabling schools and educators to adopt and deploy them locally. Specifically, we perform three tasks: segmentation of student essays into arguments, classification of the arguments by type, and assessment of their quality. We empirically evaluate the models on the “Feedback Prize – Predicting Effective Arguments” dataset of grade 6--12 students essays and demonstrate how fine-tuned small LLMs outperform baseline methods in segmenting the essays and determining the argument types while few-shot prompting yields comparable performance to that of the baselines in assessing quality. This work highlights the educational potential of small, open-source LLMs to provide real-time, personalized feedback, enhancing independent learning and writing skills while ensuring low computational cost and privacy.
\end{abstract}

\section{Introduction}
Writing well-structured essays can be challenging for students, as they require not only quality argument components but also cohesive connections between them \cite{scardamalia1987knowledge}. However, many students struggle to meet these requirements, often due to a lack of clear guidance on effective argumentation. High teacher-student ratios exacerbate this issue, limiting the individualized support that teachers can provide\footnote{\url{https://www.unesco.org/en/articles/global-report-teachers-what-you-need-know}}. To address this challenge, automatic argument mining has emerged as a promising solution, aiming to analyze the argumentative structure of essays and deliver targeted feedback to help students strengthen their arguments and overall essay structure \cite{lawrence2020argument,cabrio2018five}.

    \begin{figure*}[ht]
    \centering
    \includegraphics[width=1\textwidth]{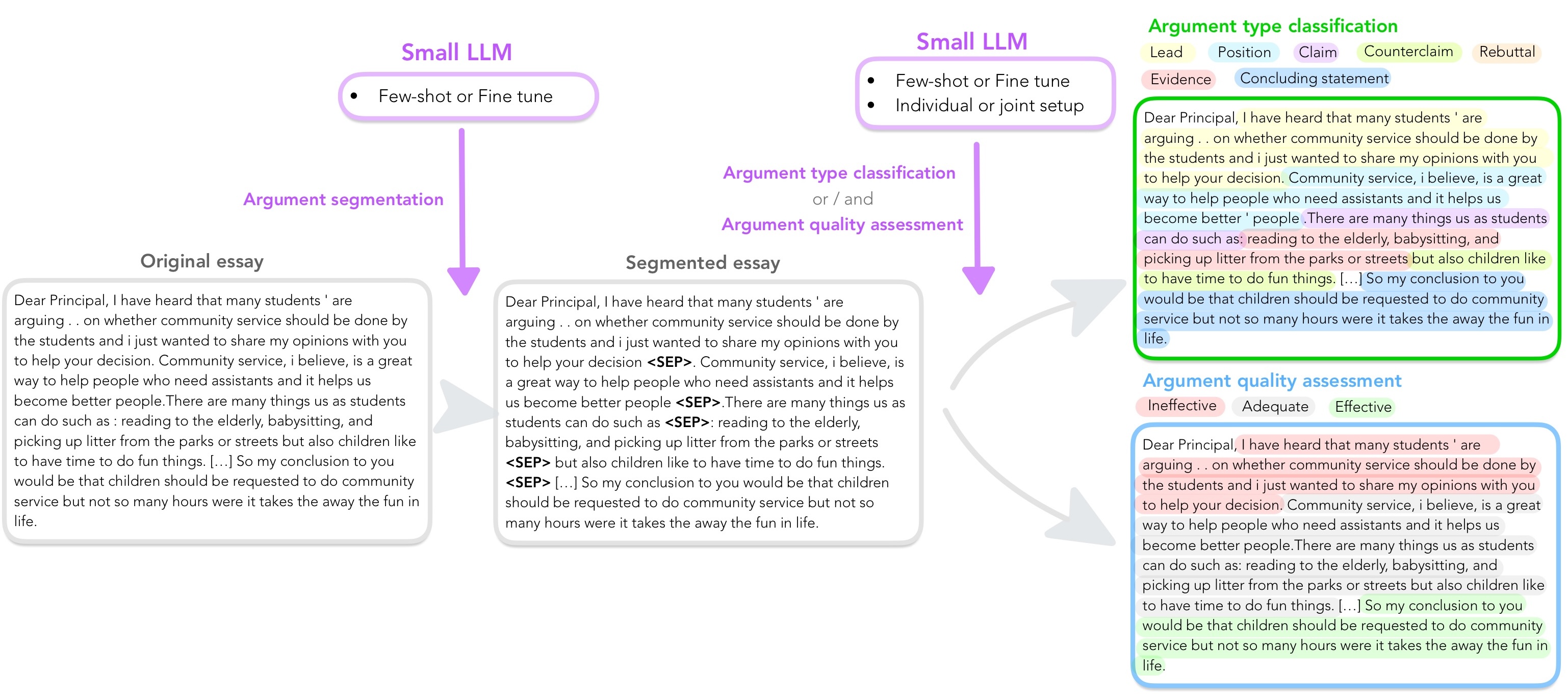}
    \caption{Overview of the proposed framework. Given an essay as input, the objective is to first segment it into arguments, then classify the argument types, and assess their quality using small open source LLMs ---Qwen 2.5 7B, Llama 3.1 8B,  and Gemma 2 9B. These tasks are performed either individually or jointly through two learning approaches: few-shot prompting or fine-tuning.}
    \label{fig:visual}
    \end{figure*}
    
Effective argument mining entails several subtasks, such as segmenting the essay into distinct argument components, classifying their type, assessing their quality and establishing relationships between them \cite{gessler2021discodisco}. These tasks can be performed sequentially or in parallel. State-of-the-art approaches in argument mining rely on encoder or encoder-decoder deep neural network-based architectures \cite{arora2023argument}. However, to date, decoder-only models remain underexplored for certain subtasks, suggesting a promising area for future exploration \cite{wachsmuth2024argument}.  

While high-performing argument mining methods have potential, they are often difficult to access and scale, especially in educational settings \cite{kashefi2023argument}. Automated Essay Scoring (AES) systems offer an alternative solution for grading essays, focusing on overall evaluation rather than on assessing individual arguments \cite{gao2024automatic}. However, limited research has explored the integration of argument mining with argument quality assessment. This work aims to bridge this gap by providing detailed feedback that helps students critically analyze their arguments and enhance their writing skills \cite{ding2023score}.

We propose leveraging open-source\footnote{We use the term open-source to refer to LLMs that are freely available with at least open-weights.}, small Large Language Models (LLMs) to perform a complete pipeline in argument mining (from segmenting the text in arguments to classifying their type and assessing their quality) in an educational context. By means of few-shot prompting and fine-tuning, these models can execute these tasks locally on the student's laptop, ensuring accessibility and maintaining computational efficiency.  
Figure~\ref{fig:visual} depicts the proposed method's pipeline, outlining each stage of the process.
    
    %Plan
    This paper is organized as follows: Section \ref{sec:litreview} reviews the most relevant literature, providing the background and context for our research. In Section \ref{sec:method}, we describe our methodology. Section \ref{sec:res} presents and analyses our experimental results.

    Finally, Section \ref{sec:discussion} provides a discussion of the findings, followed by a conclusion and an outline of the limitations in Section \ref{sec:cc} 

    Our code is available at \url{https://anonymous.4open.science/r/ACL_2025-45CE/}.
    
\section{Related work} \label{sec:litreview}
    \paragraph{Argument mining}%_____________________________
    Argument mining is a complex field that aims to identify, classify, and analyze argumentative structures within text \cite{lawrence2020argument}, drawing inspiration from frameworks, such as Toulmin's model of argumentation \cite{toulmin2003uses}. Argument mining involves numerous subtasks \cite{arora2023argument}, including argument detection, classification, assessment, and relation prediction, making end-to-end solutions particularly challenging \cite{cabrio2018five}. Thus, despite its importance, few studies tackle the full argument mining pipeline due to its complexity and methodological diversity. \cite{cao2023autoam, bao2022generative, morio2022end}.

    %\subsection{Argument mining techniques}%_____________________________    
     State-of-the-art methods in argument mining typically rely on deep neural networks \cite{arora2023argument}. Recently, advancements in Large Language Models (LLMs) have pushed the field forward. For instance, T5 \cite{raffel2020exploring} has been applied effectively to argument mining tasks \cite{kawarada2024argument}, while models like Longformer \cite{ding2023score} and BERT-based approaches \cite{kashefi2023argument} have demonstrated competitive performance across various subtasks. More complex systems provide end-to-end solutions by combining models like BART \cite{lewis2019bart} with prompting and graph-based approaches \cite{sun2024pita}, or by leveraging graph prefix tuning to enhance discourse-level understanding \cite{sun2024discourse}. Recently, \citet{gorur2024can} demonstrates that prompt-tuned, open-source models like Llama-2 \cite{touvron2023llama} and Mixtral \cite{jiang2024mixtral} can outperform state-of-the-art RoBERTa-based baselines \cite{ruiz2021transformer} in identifying agreement and disagreement relations among arguments. However, to the best of our knowledge, no research has explored to date the use of open-source, small LLMs for the combined tasks of argument classification and quality assessment. In this paper, we aim to fill this gap. 
    
   \paragraph{Educational multi-task argument mining}%___________________
   Educational multi-task argument mining focuses on extracting, classifying, and evaluating arguments in student essays—a challenging problem due to the noisy, resource-constrained nature of student writing \cite{kashefi2023argument}. Beyond the tasks of argument segmentation and classification, assessing the quality of arguments is essential for evaluating their persuasiveness and coherence \cite{wachsmuth2024argument}. Existing approaches, such as Longformer-based classification methods \cite{ding2022don} and graph-based frameworks \cite{marro2022graph}, have made contributions to this area. 

   Providing meaningful feedback from such analyses is particularly impactful in educational contexts. Actionable feedback enables students and educators to identify strengths and areas for improvement, with standardized scoring systems serving as valuable tools to guide learning and enhance outcomes \cite{cabrio2018five}. Moreover, incorporating discourse-level features has been shown to improve performance by offering deeper insights into argument structures  \cite{deshpande2023contextualizing}.

  \paragraph{Contributions} In this paper, we make several contributions to this domain. First, we address the gap in leveraging small, open-source LLMs for argument mining, combining argument segmentation, type classification, and quality assessment. Second, we propose a computationally efficient and privacy-preserving approach, enabling local analysis on standard devices through fine-tuning and few-shot prompting of the LLMs. Finally, by evaluating our approach on a benchmark dataset of student essays, we demonstrate its ability to deliver actionable feedback on a local computer, fostering improved writing skills for students grades 6-12 while preserving privacy. Our method advances argument mining in resource-constrained educational settings and highlights the transformative potential of LLMs in personalized education. 
    
\section{Method}\label{sec:method}

  \subsection{Tasks and models} 
  Given a human-written essay, the goal is to perform three tasks: (1) segmenting it into arguments; (2) classifying each argument into one of seven categories—\textit{Lead, Position, Claim, Counterclaim, Rebuttal, Evidence, Concluding Statement}; and (3) assessing the quality of each argument using three levels —\textit{Ineffective}, \textit{Adequate}, \textit{Effective}. Segmentation is conducted first, followed by argument type classification and quality assessment. These latter tasks can be performed independently or jointly.

We investigate the effectiveness of three open-source, small LLMs, namely Qwen 2.5 7B \cite{yang2024qwen2}, Llama 3.1 8B \cite{dubey2024llama}  and Gemma 2 9B \cite{team2024gemma}, for these tasks using both few-shot prompting and fine-tuning. We compare against a state-of-the-art baseline \cite{ding2022don,ding2023score} and GPT-4o mini \cite{achiam2023gpt} to shed light on the performance of open-source vs closed-source models.  
Appendix~\ref{sec:other_models} provides additional details on these models, as well as experimental results for three additional small, open-source LLMs (Llama 3.2 2B, DeepSeek R1 7B, and OLMo 2 7B).
  
    \subsection{Few-shot prompting} \label{sec:few_shot}%_______________________________
The first approach applies few-shot prompt-tuning sequentially: first, the LLM is prompted to segment the essay into arguments. Then another prompt is used for argument type classification and/or quality assessment.    
The prompts are built by concatenating the following textual elements:

    \noindent\textbf{1. Few-shot examples} are given in the format depicted by the examples included in Tables~\ref{tab:eg_segmentation}, ~\ref{tab:eg_shot_type} and ~\ref{tab:eg_shot_quality} in Appendix~\ref{sec:details_format}, depending on the task. The label of the type of argument or its quality is added between each argument in the essay. The example essays are extracted from the training set split used in \citet{ding2023score} and described in Section \ref{sec:finetune}.
            
    \noindent\textbf{2. The essay}. The segmentation task takes the original essay as input. For argument type classification and quality assessment, the essay is provided in a pre-segmented format, as illustrated in Table~\ref{tab:eg_segmentation}  in Appendix~\ref{sec:details_format}. In this format, each argument is separated by a designated \textit{SEP} label, which corresponds to the output of the segmentation task.
            
    \noindent\textbf{3. The query}, specifying the role of the LLM and the overall instructions for the task. The detailed formulations for the segmentation, argument type classification and quality assessment tasks are provided in Table~\ref{tab:queries} in Appendix~\ref{app:few-shot}.
            
   \noindent\textbf{4. Output requirements.} For the argument type classification and argument quality assessment tasks, the model is asked to generate the output in a specific JSON format, described in the prompt as follows:
            \textit{“For the given} argument component, \textit{identify its [type] and/or [quality]. Provide the output as a JSON object with the key: [TYPE] and/or [QUALITY].”}
            For segmentation, the output is described in the prompt as follow: "\textit{Place <SEP> immediately at the end of each segment. Preserve all original words, spacing, and order.}"

    \noindent\textbf{5. The specific argument} of the essay that the LLM is asked to analyze in the argument type classification and argument quality assessment tasks.

Examples of the expected output format and the handling of incorrect outputs are detailed in Appendix~\ref{sec:detail_exp}.
    %comment on the choice of the prompt
    After segmentation, classifying the entire text at once may cause label mismatches, misaligning them with arguments. To simplify this task for smaller LLMs, we classify one argument per prompt. However, the full essay is still provided as context, given that the type classification and quality assessment of each argument depend on the essay's overall argumentation strategy.

    Also note that the specific structure, content and phrasing of each prompt have been meticulously designed after an intensive effort of prompt engineering. Any changes to the sequence or omission of parts significantly affect both the accuracy and the format of the output.
    %alternative approaches:
    An alternative approach to prompt-tuning the LLM would involve asking it to replace each separator with its corresponding label (type, quality, or both), as demonstrated in the fine-tuning process described in the next section. With this method, the output should exactly replicate the original essay, with classified labels inserted at the appropriate points between arguments. However, this approach did not deliver good performance with the small LLMs as these models struggle to reproduce the input text word-for-word and often introduce errors such as adding extra words, omitting parts of the text, or inserting additional separators. For output examples, see Appendix \ref{sec:details_format}. 

    \subsection{Fine-tuning}\label{sec:finetune}
    The second approach involves fine-tuning an LLM for the following tasks: (1) argument segmentation; (2) argument type classification; (3) argument quality assessment; and (4) argument type classification and quality assessment together. The three small, open-source models were fine-tuned on Google Colab using an A100 GPU on each of the tasks. To establish a comparative upper bound, we also fine-tuned GPT-4o mini using OpenAI's API. Further details on the fine-tuning of this model can be found in Appendix \ref{sec:FT_GPT}.
    
    \paragraph{Data} Fine-tuning requires the use of a dataset. In our study, we used a collection of argumentative essays written by U.S. students in grades 6-12, annotated by expert raters. The dataset is derived from the Kaggle competition “Feedback Prize - Predicting Effective Arguments,” \footnote{\url{https://www.kaggle.com/competitions/feedback-prize-effectiveness/data}} which constitutes a subset of the \textit{PERSUADE 2.0 Corpus}. This subset includes approximately 6,900 essays from a total of 26,000 argument components, representing just over a quarter of the corpus. The essays were selected to achieve a balanced distribution of discourse elements across varying levels of effectiveness \cite{crossley2022persuasive,crossley2023large}. The dataset encompasses the seven argument types derived from Toulmin’s argumentation model \cite{toulmin2003uses}: \textit{Lead, Position, Claim, Counterclaim, Rebuttal, Evidence,} and \textit{Concluding Statement}; a quality assessment for each argument is also provided: \textit{Ineffective}, \textit{Adequate}, or \textit{Effective}.

    For fine-tuning, we used the split provided by \citet{ding2023score} consisting of 3,353 essays (\emph{i.e}, 29,440 arguments) for the training set and 419 essays (\emph{i.e.}, 3,614 arguments) for the validation set. Detailed statistics of this dataset split can be found in \citet{ding2022don,ding2023score}. 
    
    \paragraph{Setting}
    We performed the fine-tuning of the small, open-source LLMs using the \textit{SFTTrainer} module from the \textit{TRL} library\footnote{\url{https://huggingface.co/docs/trl/sft_trainer}}. We employed \textit{Unsloth}\footnote{\url{https://github.com/unslothai/unsloth}} to optimize performance and reduce memory usage.

    Additionally, we incorporated Low-Rank Adaptation (LoRA) \cite{hu2021lora} and Quantized Low-Rank Adaptation (QLoRA) \cite{dettmers2024qlora} to further reduce memory demands and improve the fine-tuning speed. An early stopping criterion was applied to optimize performance and reduce memory usage. The specific hyperparameters used in the fine-tuning can be found in Appendix  \ref{sec:hyper}. Regarding GPT 4o-mini, we used OpenAI's API for fine-tuning this model. 
 
    \paragraph{Input sequence and target}
    Both the input sequence and the fine-tuning target were formatted in accordance with the previously described few-shot prompting methodology. For the segmentation task, the input was the original essay, while the target output was the corresponding essay segmented by a designated \textit{SEP} label (See Table~\ref{tab:eg_segmentation}, in Appendix). 
    In contrast, for the joint task of argument type classification and quality assessment, the input consisted of the essay already segmented with the \textit{SEP} labels (See Table~\ref{tab:eg_segmentation}, in Appendix). The target, in this instance, was defined as the same essay further partitioned into discrete arguments, with each argument interleaved with labels specifying both the type of argument and its quality (See Table~\ref{tab:eg_shot_together}, in Appendix). This approach ensures methodological consistency across tasks and facilitates a systematic evaluation of model performance on both segmentation and combined argument type classification and quality assessment tasks. See Appendix~ \ref{sec:ft_details} for more details.

    \paragraph{Inference}
    The fine-tuned models were run on an Apple M1 Pro laptop with 32 GB RAM using Ollama\footnote{\url{https://github.com/ollama/ollama}, \url{https://ollama.com}}, an open-source framework that enables users to run, create, and share LLMs locally on their machines. We did not provide any few-shot examples demonstrating how to perform the task or how to specify the output format. To evaluate performance, we used the same test set as \citet{ding2023score}. Fine-tuning is expected to enhance the model's ability to generate outputs that closely mirror the input essay and conform better to the specified output format than the non-fine-tuned models. See Table~\ref{tab:exp_variant} in Appendix, for a summary of experiment variants.

\section{Evaluation}\label{sec:res}            
     \subsection{Dataset}
     We performed all our evaluations on the test set of the “Feedback Prize - Predicting Effective Arguments,” \footnote{\url{https://www.kaggle.com/competitions/feedback-prize-effectiveness/data}} datasets. 
       We use the same test set employed by \citet{ding2023score}, consisting of 419 essays with a total of 3,711 arguments. Detailed statistics of this dataset split can be found in \citet{ding2022don,ding2023score}.
For the segmentation task,  the essay has to be segmented into arguments. For the argument type classification task, the segmented arguments need to be classified into one of seven types: \textit{Lead, Position, Claim, Counterclaim, Rebuttal, Evidence,} and \textit{Concluding Statement}. In the case of the quality assessment task, the possible values are: \textit{Ineffective}, \textit{Adequate, Effective}. See Table~\ref{tab:queries} in the Appendix~\ref{app:few-shot} for a description of each label.

     \subsection{Performance metrics} \label{sec:setup}

     To assess the efficiency of the proposed methods, we report the following metrics that consider the imbalance in the distribution of labels.

\textbf{Metrics per label}, namely precision, recall, and F1 score for each individual label to have an in-depth look at the classifier’s performance on a per-label basis.    

\textbf{Multi-label confusion matrix}
   to provide a detailed breakdown of model performance across argument categories, highlighting both accurately predicted cases (along the diagonal) and common misclassifications. The values in the matrix give insights into which argument types or quality assessment labels the proposed method distinguishes effectively and where it struggles.
    
\textbf{Macro-averaged F1 score}, 
    which is the mean of the F1 scores for each label, treating all labels equally, thereby providing a measure of overall performance across all labels without considering label imbalance \footnote{See the formula in Appendix \ref{sec:metric_details}}. This metric is a standard in the argument mining community \cite{bao2022generative,sun2024pita,ding2023score,morio2022end}.

\textbf{Segmentation F1 score}
We compute the F1 score for segmentation at the token level using the BIO framework, where each token at the \textit{beginning} of an argument is tagged as \textit{B}, tokens \textit{inside} the argument are tagged as \textit{I} and \textit{O} denotes when a token is not part of an argument, which is not the case in our task as all the tokens are supposed to belong to an argument.

    \textbf{Type and quality F1 score}
    We adopt the evaluation method used by \citet{ding2023score}. A predicted argument ($S_p$) with at least 50\% of overlap with a gold argument ($S_g$) is considered a \emph{match} i.e: $\min(o_{gold},o_{pred}) > 0.5$, where: $o_{gold} = \frac{|S_g| \cap |S_p|}{|S_g|}$ is the overlap of the predicted argument with the gold argument and $o_{pred} = \frac{|S_g| \cap |S_p|}{|S_p|}$ is the overlap of the gold argument with the predicted argument. 
    Matched predicted arguments are considered \emph{true positives} if they are of the same type (or quality) as the gold argument. Otherwise, they are classified as a false negative. Unmatched predicted arguments are considered false positives and labeled as \textit{Echec}.

    \textbf{Spelling errors} LLMs are required to reproduce each input essay—originally written by children—segmented with the \textit{SEP} separator. Because these essays contain numerous spelling errors, smaller LLMs often attempt to correct them automatically, thereby altering the text and complicating direct comparisons 
    %the overlap 
    with the ground truth. Since our goal is not to address spelling errors and the dataset labels do not account for them, we first correct the essays before providing them to the LLM to ensure accurate and consistent segmentation. To do so we use \texttt{language-tool-python}, a wrapper for LanguageTool\footnote{\url{https://languagetool.org}}.
    
    \subsection{Baselines} \label{sec:baselines}
    We compare our method with several encoder-based variants following the approach proposed by \citet{ding2023score}. Their framework employs BERT \cite{devlin2018bert} for argument type and quality assessment and a Longformer \cite{beltagy2020longformer} for token-level segmentation, leveraging the Longformer’s ability to handle long-text classification (see Appendix~\ref{sec:encoder}). Notably, their method is the only approach in the literature that utilizes this dataset for the same tasks while processing segmented essays as input. Additionally, as mentioned earlier, we compare our results with GPT-4o mini \cite{achiam2023gpt}, both vanilla and fine-tuned (using the same fine-tuning data as with the other small LLMs, see Appendix~\ref{sec:FT_GPT}) versions, to have a sense of the upper performance bound achieved by a commercial model.  

\subsection{Results}
    
        \subsubsection{Argument segmentation}

\begin{figure}[h!]
    \centering
    \includegraphics[width=0.45\textwidth]{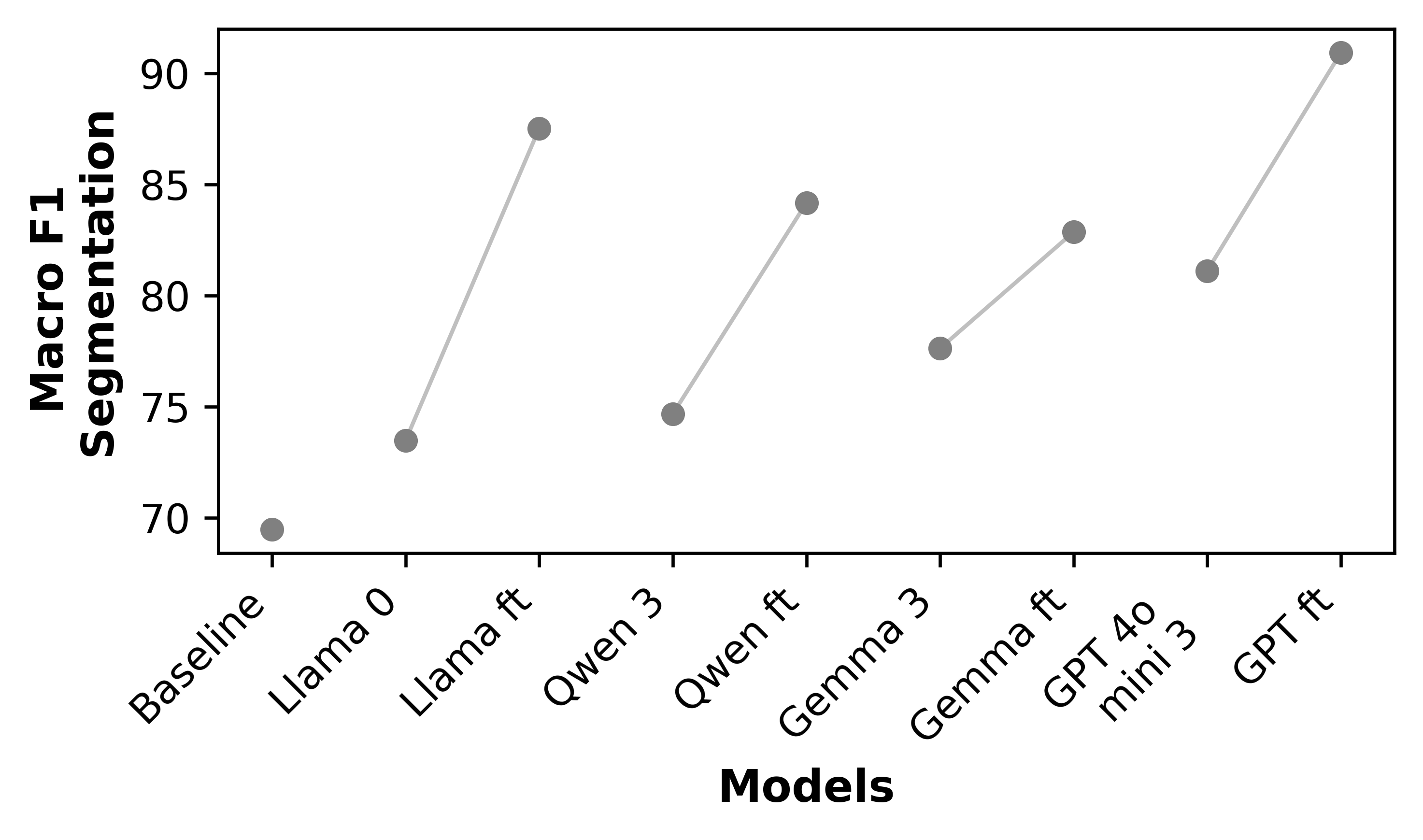}
    \caption{\textbf{Macro-averaged F1 scores [\%] for the argument segmentation task across models.} Comparison of small open-source models (Qwen 2.5 7B, Llama 3.1 8B, Gemma 2 9B) in the best few-shot (zero or three-shot) and fine-tuned (ft) settings with the baseline (Longformer) and GPT-4o mini, both with three-shot and fine-tuned. Error bars depict the standard deviation.}
    \label{fig:res_S}
    \end{figure}

    Figure~\ref{fig:res_S} compares the performance of the small, open-source models on the segmentation task, evaluated in their best few-shot settings (either zero or three) and fine-tuned configurations. Additionally, we include the Longformer as a state-of-the-art baseline, and GPT-4o mini (with three-shot learning and fine-tuned) as a commercial upper bound. Error bars indicate the standard deviation across three runs. All LLM models surpass the Longformer's performance, demonstrating the advantages of large-scale pretraining and transfer learning. Fine-tuned models exhibit substantial performance gains over their few-shot counterparts, underscoring the effectiveness of supervised adaptation. Among all small open-source models, the fine-tuned Llama 3.1 8B achieves the highest F1 score of \textbf{87.52}, an increase of \textbf{18.05} points or \textbf{26.00\%} in segmentation performance over the baseline (Longformer) and \textbf{6.41 }points but \textbf{3.7\%} below GPT-4o mini's performance. Figure~\ref{fig:res_overlap}, in Appendix, presents the  overlap of the inferred segmentation with the gold segmentation across models. Fine-tuned Llama 3.1 8B achieves the highest overlap with the ground truth, outperforming Longformer and aligning with the macro F1 results. Figure~\ref{fig:res_arg} in the Appendix~\ref{sec:segm_ana} reports the average number of arguments per essay across models for further segmentation analysis.

    \begin{figure*}[h!]
    \centering
    \includegraphics[width=0.49\textwidth]{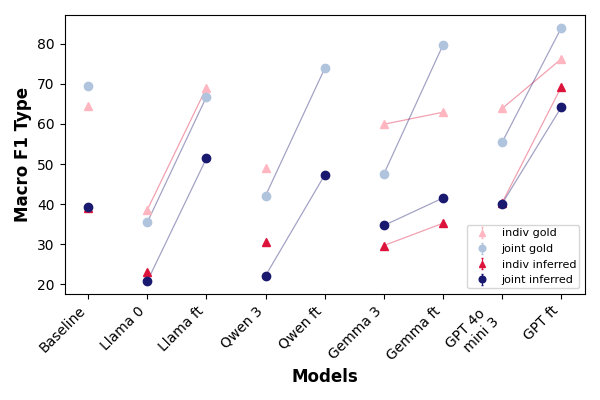}
    \includegraphics[width=0.49\textwidth]{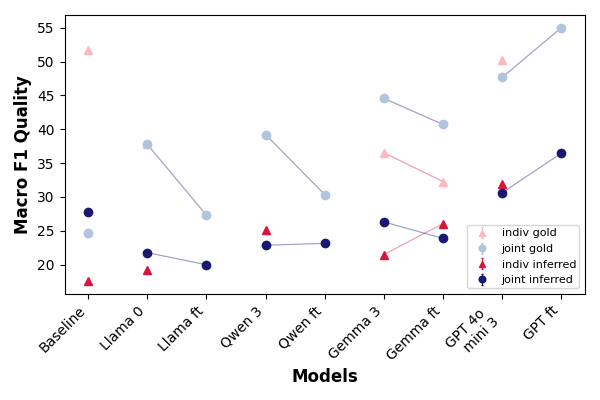}
     \caption{\textbf{Macro-averaged F1 scores [ \% ] for the argument type classification (left) and quality assessment (right) across models.} Comparison of three small open-source models (Qwen 2.5 7B, Llama 3.1 8B, Gemma 2 9B) in the best few-shot (zero or three-shot) and fine-tuned (ft) settings with the baseline and GPT-4o mini (few-shot and fine-tuned). The results highlighted in transparent colors correspond to the evaluation with the gold segmentation whereas the darker colors correspond to inferred segmentation. In the case of gold segmentation, the baseline corresponds to a BERT model with two prediction heads. In the case of inferred segmentation, the segmentation is carried out by a Longformer followed by a classification with BERT. Circles represent the joint setup (both type and quality classification are performed at the same time) whereas triangles correspond to the individual setup (type and quality classification are performed separately). Error bars show the standard deviation.}
    \label{fig:res_T}
    \end{figure*}
    
    \subsubsection{Argument type and quality assessment}
    Figure~\ref{fig:res_T} shows the macro-averaged F1 scores for the argument type (left) and quality (right) tasks, respectively, across various models and experimental settings. The figure compares the performance of the three small open-source models in the best few-shot (either zero or three) and fine-tuned settings, along with the baseline and GPT 4o mini (three-shot and fine-tuned). The F1 scores are reported for two task setups: \textit{individual} classification (indiv.) and \textit{joint} classification of both argument type and quality together (joint). Results are shown for two segmentation conditions: \textit{gold} (provided) and \textit{inferred}. Transparent colors indicate the gold condition, while less transparent colors represent inferred segmentation. Circles and triangles correspond to the joint and individual setups, respectively. Error bars represent the standard deviation of the results.

\paragraph{Argument type classification}
    Regarding the type classification task, models using the gold segmentation consistently outperform the models with inferred segmentation and the baseline. We also observe significant improvements when fine-tuning the small, open-source LLMs, both with the gold and inferred segmentation.
    The best-performing model with the gold segmentation is Gemma 2 9B fine-tuned, with an F1 score of \textbf{79.74}, which is \textbf{10.27} points or  \textbf{14.78\%} larger than the baseline (BERT) but \textbf{5.1\%} below GPT-4o mini's performance.  The best-performing model with inferred segmentation is Llama 3.1 8B fine-tuned, with an F1 score of \textbf{51.45}, which is \textbf{12.27} points or \textbf{31.32\% } larger than the baseline (Longformer + BERT) but \textbf{19.85\%} lower than GPT-4o mini fine-tuned. Generally, models yield better performance in the joint than in the individual setup. 

\paragraph{Argument quality assessment}
    Regarding the argument quality assessment task, models using the gold segmentation also consistently outperform the models with inferred segmentation. In this case, the best-performing small open-source model is Gemma 2 9B with three-shot learning, achieving an F1 score of \textbf{44.56}, which is \textbf{19.96} points or \textbf{81.14\%} larger than the baseline's performance but \textbf{18.98\%} lower than the GPT-4o mini fine-tuned performance. With inferred segmentation, the best performing small, open-source LLM (Gemma 2 9B with three-shot learning) achieves similar performance to the baseline, both of them below that of GPT-4o mini. Interestingly, fine-tuning, in this case, tends to worsen the performance across models, particularly when given the gold segmentation. 
    
    Note that the results for Llama 3.1 Qwen 2.5 and GPT-4o mini fine-tuned are absent in the individual setup due to repeated inference failures. Refer to Section~\ref{sec:lim} for more details.

    \subsubsection{Label-level performance analysis}
        To better understand the previously reported performance figures, we summarize the results at the label level for the best-performing model (Llama 3.1 8B fine-tuned in the argument type classification task and Gemma 2 9B three-shot learning in the argument quality assessment task) in the joint set up and the inferred segmentation configuration. Tables ~\ref{tab:report_ft_together_type}, ~\ref{tab:report_ft_together_eff}, (and ~\ref{tab:report_ft_together_segm} in Appendix)  depict the precision, recall and F1 score for the argument type classification, quality assesment and segmentation tasks, respectively. Furthermore, Tables~\ref{tab:CM_ft_together_type}, ~\ref{tab:CM_ft_together_eff} and ~\ref{tab:CM_ft_together_segm} in Appendix contain the corresponding confusion matrices. 
        \begin{table}[ht]
    \centering
    \setlength{\tabcolsep}{5pt}
    \resizebox{0.35\textwidth}{!}{%
    \begin{tabular}{|l|c|c|c|}
    \hline
    \rowcolor{white!30}  \textbf{Type}& \textbf{Precision} & \textbf{Recall} & \textbf{F1-score} \\ \hline
    Lead& \cellcolor[HTML]{73c476}48.48& \cellcolor[HTML]{00441b}\textcolor{white}{96.97}& \cellcolor[HTML]{36a055}64.58\\ \hline
    Position& \cellcolor[HTML]{63bc6e}52.63& \cellcolor[HTML]{47ae60}58.94& \cellcolor[HTML]{56b567}55.59\\ \hline
    Claim& \cellcolor[HTML]{90d18d}{41.17}& \cellcolor[HTML]{2a924a}69.86& \cellcolor[HTML]{65bd6f}51.75\\ \hline
    C claim& \cellcolor[HTML]{74c477}48.34& \cellcolor[HTML]{5eb96b}53.70& \cellcolor[HTML]{6cc072}50.71\\ \hline
    Rebuttal& \cellcolor[HTML]{65bd6f}51.67& \cellcolor[HTML]{79c679}47.09& \cellcolor[HTML]{78c679}47.49\\ \hline
    Evidence& \cellcolor[HTML]{5ab769}54.91& \cellcolor[HTML]{00682a}\textcolor{white}{86.33}& \cellcolor[HTML]{2f9951}67.11\\ \hline
    Concluding& \cellcolor[HTML]{137e3b}\textcolor{white}{77.92}& \cellcolor[HTML]{278e48}\textcolor{white}{71.28}& \cellcolor[HTML]{1d8742}\textcolor{white}{74.36}\\ \hline
    \end{tabular}}
    
    \caption{Performance evaluation (precision, recall and F1-score per label) in the argument type classification task using the fine-tuned Llama 3.1 8B model on the joint setup with inferred argument segmentation.}
    \label{tab:report_ft_together_type}
    \end{table}
            
\paragraph{Argument type classification} As seen in Table \ref{tab:report_ft_together_type} the model is the most accurate in classifying \textit{Concluding statements}, followed by \textit{Evidence}. In contrast, the model exhibits the lowest performance when classifying \textit{Rebuttals} and \textit{Counterclaims}. Interestingly, the model has very high recall (96.97\%) but low precision (48.48\%) when classifying \textit{Lead} arguments, suggesting that it often over‐labels arguments as \textit{Lead}. There is large misclassification rates of \textit{Claim}, \textit{Position}, and \textit{Evidence}, into \textit{Echec} entries, confirming that segmentation errors contribute to lowering the performance in the classification task, see Table~\ref{tab:CM_ft_together_type} in the Appendix.

        \paragraph{Quality assessment} Regarding quality assessment, the best performance of the model is in the \textit{Adequate} label, struggling with the \textit{Ineffective} (low recall) and especially the \textit{Effective} (low precision) labels. In fact, \textit{Adequate} tends to be over-predicted by the model. Part of the classification errors are also due to segmentation mistakes, as reflected by the  \textit{Echec} column in the confusion table in Table~\ref{tab:CM_ft_together_eff} in the Appendix. 
    
            %report_ft_together_eff
        \begin{table}[h!]
        \centering
        \setlength{\tabcolsep}{5pt}
        \resizebox{0.32\textwidth}{!}{%
        \begin{tabular}{|l|c|c|c|}
        \hline
        \rowcolor{white!30} \textbf{Quality} & \textbf{Precision} & \textbf{Recall} & \textbf{F1-score} \\ \hline
    Ineffective& \cellcolor[HTML]{3686c0}42.00& \cellcolor[HTML]{75b4d8}29.78& \cellcolor[HTML]{58a1cf}34.84\\ \hline
Adequate& \cellcolor[HTML]{549ecd}35.98& \cellcolor[HTML]{09306b}\textcolor{white}{62.74}& \cellcolor[HTML]{2676b8}45.70\\ \hline
Effective& \cellcolor[HTML]{c4daee}15.86& \cellcolor[HTML]{09539e}\textcolor{white}{54.34}& \cellcolor[HTML]{97c6df}24.65\\ \hline

        \end{tabular}}
        
       \caption{Performance evaluation per label in the argument quality assessment task using Gemma 2 9B three-shot model on the joint setup and inferred segmentation.}
       \label{tab:report_ft_together_eff}
        \end{table}    
\section{Discussion}\label{sec:discussion}

In this paper, we have explored the potential of three small, open-source LLMs---namely, Qwen 2.5 7B, Llama 3.1 8B, Gemma 2 9B---to perform three argument mining tasks in an educational setting: argument segmentation, argument type classification, and argument quality assessment. We have experimented with both few-show prompting and fine-tuning, comparing small open-source LLMs to commercial LLMs (GPT-4o mini) and state-of-the-art encoders. From these extensive experiments, we draw several findings. 

First, small and open-source LLMs are able to effectively perform argument mining tasks with significantly better performance than state-of-the-art baselines (Longformer and BERT). Commercial small LLMs, such as GPT4o-mini, yield the best performance.

Second, different approaches provide the best results depending on the task. Whereas fine-tuned models consistently outperform few-shot prompting in the argument segmentation and type classification tasks, we observe the opposite behavior in the argument quality assessment task. 

Third, model performance in the tasks of interest does not necessarily increase with the number of shots when performing few-shot prompting, which is consistent with what has been reported in the literature \cite{liu2024lost}. This finding is particularly evident in the case of Llama 3.1 8B, which exhibits the best performance in tasks with zero-shot prompting. The deterioration in performance with the number of shots is probably due to the complexity of longer prompts which seemed too hard for the model to make sense of. 

Fourth, joint fine-tuning setups where both tasks were carried out at the same time tend to yield better results than individual setups where the tasks were performed independently, showing a strong link between classifying argument types and assessing their quality \cite{crossley2023large}. 

Fifth, the automatic segmentation and classification of the type of argument seems to be an easier task than the assessment of the quality of the arguments, likely due to difficulties in creating high-quality and consistent ground truth quality assessments across essays \footnote{See Appendix~\ref{sec:qual_future} for a discussion on the annotation quality} \cite{wachsmuth2024argument}. 

Finally and most importantly, this study highlights the potential of open-source, small LLMs, running locally on personal computers, to support students in the development of their essay-writing skills. Our framework prioritizes privacy and accessibility, addressing the challenge of developing efficient models for local use without consuming too many resources \cite{kashefi2023argument}.

\section{Conclusion and future work} \label{sec:cc}

In this paper, we have presented a study of the potential of small, open-source LLMs for argument mining, investigating their effectiveness in both few-shot prompting and fine-tuning setups. Fine-tuning proved especially valuable for argument segmentation and type classification such that small, open-source LLMs significantely outperform state-of-the-art approaches by 18.05 points in argument segmentation and 10.27 points in argument type classification. Our experiments also illustrate the value of joint setups for improved argument type classification. By focusing on models running locally on students' personal computers, our research promotes accessibility and resource efficiency, illustrating the potential of open-source, small LLMs as a promising tool for educational applications. 

\section*{Limitations} \label{sec:lim}

Our work is not exempt from limitations that we plan to address in future work. First, the performance on the argument quality assessment task is low for all models, including state-of-the-art methods. We hypothesize that the poor performance might be due to the quality of the annotated data, as suggested by other authors \cite{crossley2022persuasive,crossley2023large}. Hence, we plan to improve the annotation quality and create a more reliable dataset to be shared with the research community. 

Second, the fine-tuned Qwen 2.5 7B, Llama 3.1 8B, and GPT4o mini in the individual setup were unable to perform the argument type classification and quality assessment tasks due to repeated failures during inference. While the exact cause remains to be investigated, potential reasons include the hyperparameters used for the fine-tuning not being suitable for the individual setup or instability in the fine-tuned models.

Finally, we evaluated small LLMs using the \textit{PERSUADE 2.0 corpus} dataset, which consists solely of English high-school writing. As a result, it remains uncertain whether findings can be generalized to other educational contexts and languages, which could be explored in future work. 

\section*{Ethical considerations}
The integration of AI-based argument mining for automatically evaluating student essays raises ethical concerns related to data privacy, fairness, and accountability. First, sensitive educational data, which is often tied to the students’ identities, must be handled securely, with clear consent and transparent data-sharing policies. Second, the presence of biases in the annotation and model training processes can lead to inequitable outcomes that disadvantage specific student groups. Third, blind trust and over-reliance on automated assessments can lower the teachers’ professional judgment and lead to a lack of human touch in the educational process. The proposed approach is part of a larger project aiming to develop an educational chatbot that mitigates these concerns by leveraging small, open-source LLMs that run locally on the students' computers and by consciously measuring and mitigating biases both in the training data and the models. 
\section*{Acknowledgements}
L.F. and N.O. have  been partially supported by a nominal grant received at the ELLIS Unit Alicante Foundation from the Regional Government of Valencia in Spain (Resolución de la Conselleria de Innovación, Industria, Comercio y Turismo, Dirección General de Innovación). L.F. has also been partially funded by a grant from the Banc Sabadell Foundation.

%\bibliography{latex/custom}

\appendix
 
\section{Appendix}
\subsection{Detailed experimental setup and methodology}\label{sec:exp}
The experimental setup consists of multiple configurations based on different task types, segmentation methods, adaptation strategies, and model choices. Table~\ref{tab:exp_variant} provides an overview of the experiment variants. The tasks include argument segmentation, argument type classification, and argument quality assessment. Two segmentation approaches are considered: gold segmentation (ground truth) and inferred segmentation (automatically performed by the model). The setup can be either individual, where argument type and quality are classified separately, or joint, where both are assessed together. We explore two adaptation strategies: few-shot learning (ranging from zero to four-shot) and fine-tuning. The models used in the experiments fall into three categories: encoder-based models, small open-source LLMs, and a proprietary model, GPT-4o mini. 

\begin{table}[h!]
    \centering
    \renewcommand{\arraystretch}{1.1} 
    \resizebox{0.47\textwidth}{!}{%
    \begin{tabular}{|c| l|} 
                    \Xhline{2\arrayrulewidth} % Thicker first horizontal line
        \textbf{Category} & \textbf{Options} \\             \Xhline{2\arrayrulewidth} % Thicker first horizontal line
        \multirow{3}{*}{Task} 
         & - Argument segmentation \\
         & - Argument type classification \\
         & - Argument quality assessment \\ \hline
        \multirow{2}{*}{Segmentation} 
         & - Gold \\
         & - Inferred \\ \hline
        \multirow{2}{*}{Setup} 
         & - Individual: type or quality \\
         & - Joint: type and quality\\ \hline
        \multirow{2}{*}{\makecell{Adaptation \\ strategy} } 
         & - Few-shot (zero to four-shot) \\
         & - Fine-tuned \\ \hline
                \multirow{1}{*}{Models} & - Encoder-based: \\
                & \hspace{0.4cm}  - BERT\\
                & \hspace{0.4cm}  - BERT with two heads\\
                &  \hspace{0.4cm} - Longformer\\
                &  \hspace{0.4cm} - Longformer with two heads\\
                & - Small open source LLMs: \\ 
                    &  \hspace{0.4cm} - Llama 3.2 3B\\
                    &  \hspace{0.4cm} -  OLMo 2 7B \\
                    &  \hspace{0.4cm} - Qwen 2.5 7B \\ 
                    &  \hspace{0.4cm} - DeepSeek R1 7B \\ 
                    &  \hspace{0.4cm} - Llama 3.1 8B \\
                    &  \hspace{0.4cm}- Gemma 2 9B \\
                & - GPT 4-o mini \\      
                \Xhline{2\arrayrulewidth} % Thicker first horizontal line
                
    \end{tabular}}
    \caption{ Summary of experiment variants: detailing different tasks, segmentation methods, experimental setups, adaptation strategies, and the models used.}
    \label{tab:exp_variant}
\end{table}

\subsection{Further details on adaptation strategies}
\subsubsection{Encoder-based baseline}\label{sec:encoder}  
The following encoders are used for different task variants, following the approach proposed by \citet{ding2023score}:  
\begin{itemize}  
    \item \textbf{BERT} \cite{devlin2018bert} for argument type classification and argument quality assessment (individual setup) with segmentation given.  
    \item \textbf{BERT with two prediction heads} \cite{ding2023score} for joint argument type classification and argument quality assessment (joint setup) with segmentation given.  
    \item \textbf{Longformer} \cite{beltagy2020longformer} for segmentation via token classification.  
    \item \textbf{Longformer with two prediction heads} \cite{ding2023score} for segmentation and argument type classification and segmentation and argument quality assessment (joint setup).  
\end{itemize}
The training was conducted for 10 epochs and using the same specific setting used in \citet{ding2023score}. The evaluation follows the same methodology used to assess the LLMs’ performance.

\subsubsection{Few-shot learning}\label{app:detail_few_shot}
\paragraph{Few-shot prompting queries}\label{app:few-shot}
Complementing the Section \ref{sec:few_shot} in the main paper, Table~\ref{tab:prompt_tuning_outputs} presents the expected output format for different tasks, while Table~\ref{tab:queries} lists the queries used to prompt the LLMs.
        
\begin{table}[h!]
    \centering
    \renewcommand{\arraystretch}{1.2} 
    \setlength{\tabcolsep}{5pt}
    \resizebox{0.4\textwidth}{!}{%
        \begin{tabular}{!{\vrule width 1.5pt}c|c!{\vrule width 1.5pt}}
            \Xhline{2\arrayrulewidth} % Thicker first horizontal line
            \textbf{Tasks} &  \textbf{Few-shot output format} \\
            \Xhline{2\arrayrulewidth} % Thicker separator
            \textbf{Type} & \{"TYPE": ["Position"]\} \\ 
                        \hline
            \textbf{Quality} & \{"QUALITY": ["Adequate"]\} \\ 
            \hline
            \textbf{Type and Quality} & \makecell[l]{\{"TYPE AND QUALITY": \\ \quad ["Position", "Adequate"]\}} \\ 
            \Xhline{2\arrayrulewidth} % Thicker bottom line
        \end{tabular}
    }
\caption{Expected output format for few-shot learning for the argument type classification and quality assessment task in the individual and joint setups.}  
    \label{tab:prompt_tuning_outputs}  
\end{table}

\begin{table*}[h!]  
\centering  
\resizebox{0.91\textwidth}{!}{%
\begin{tabular}{|p{2.5cm} p{12cm}|}
\hline
\textbf{Task} & \textbf{Instructions} \\ [0.5em] 
\hline
\textbf{Segmentation} & 
\#TASK: Segment the following essay into distinct argument components.  
After each argument component, insert the marker \texttt{<SEP>}.  
Keep the original text in the same order without adding, removing, or altering any words (other than inserting the \texttt{<SEP>} markers).\\ &  
\#GUIDELINES: Identify each coherent segment that forms a logical unit of the argument (e.g., claims, premises, evidence, or conclusions). \\ [1em] 
\hline
\textbf{Type} & 
You are a strict AI evaluator specializing in detecting the type of argument components in essays.  
The argument types are as follows:  

- \textit{Lead}: An introduction that begins with a statistic, quotation, description, or other device to grab the reader’s attention and point toward the thesis.\\& 
- \textit{Position}: An opinion or conclusion on the main question.\\&  
- \textit{Claim}: A statement that supports the position.\\&  
- \textit{Counterclaim}: A statement that opposes another claim or provides an opposing reason to the position.\\&  
- \textit{Rebuttal}: A statement that refutes a counterclaim.\\&  
- \textit{Evidence}: Ideas or examples that support claims, counterclaims, or rebuttals.\\&  
- \textit{Concluding Statement}: A statement that restates the claims and summarizes the argument. \\ [1em] 
\hline
\textbf{Quality} & 
You are a strict AI evaluator specializing in assessing the quality of argument components in essays.  
Each component should be rated as one of the following:  

- \textit{Ineffective}: The component is unclear, unconvincing, or poorly structured.\\&  
- \textit{Adequate}: The component is understandable and somewhat convincing but lacks strong support or clarity.\\&  
- \textit{Effective}: The component is well-structured, clear, and strongly supports the argument. \\ [0.5em] 
\hline
\end{tabular}}
\caption{Queries used in LLM few-shot prompting (Section \ref{sec:few_shot}) for segmentation, argument type classification, and quality assessment tasks.} 
\label{tab:queries}  
\end{table*}

\subsubsection{Fine-tuning}\label{sec:ft_details}
Following the dataset split used in \cite{ding2023score}, we fine-tuned the small LLMs and GPT-4o-mini on a training set of 3,353 essays (29,440 argument segments) and evaluated them on a validation set of 419 essays (3,614 argument segments).

\paragraph{Input sequences and target formats}
Complementing the Section \ref{sec:finetune} in the main paper, 
Table~\ref{tab:task_formats} provides the input sequences and target formats used during the fine-tuning for the different tasks. See Tables~\ref{tab:eg_segmentation},~\ref{tab:eg_shot_type}, ~\ref{tab:eg_shot_quality} and ~\ref{tab:eg_shot_together}  in Appendix \ref{sec:details_format} for examples of these formats.
\begin{table}[h!]
    \centering
    \renewcommand{\arraystretch}{1.1} 
    \setlength{\tabcolsep}{3pt}
    \resizebox{\columnwidth}{!}{ 
        \begin{tabular}{!{\vrule width 1.5pt}c|c|c!{\vrule width 1.5pt}}
            \Xhline{2\arrayrulewidth} 
            \textbf{Tasks} & \textbf{Input Format} & \textbf{Target Format} \\ 
            \Xhline{2\arrayrulewidth} % Thicker separator below headers
            \textbf{Segmentation} & Essay & Essay + \textit{SEP} separator\\ 
            \hline
            \textbf{Type (Indiv.)} & Essay + \textit{SEP} & Essay + \textit{TYPE} separators \\ 
            \hline
            \textbf{Quality (Indiv.)} & Essay + \textit{SEP} & Essay + \textit{QUALITY} separators \\ 
            \hline
            \textbf{Type + Quality} & Essay + \textit{SEP} & Essay + \textit{TYPE} and \textit{QUALITY} sep. \\ 
            \Xhline{2\arrayrulewidth} % Thicker bottom line
        \end{tabular}
    }
    \caption{Input sequences and target formats for different tasks when fine-tuning the models.}
    \label{tab:task_formats}
\end{table}

\paragraph{Fine-tuning small, open source LLMs} \label{sec:hyper}
        Qwen 2.5 7B, Llama 3.1 8B, and Gemma 2 9B were fine-tuned on a single GPU. We used the AdamW optimizer ($\beta_1 = 0.9$, $\beta_2 = 0.999$) in 8-bit precision. A weight decay of $0.01$ was applied to all weights except biases and normalization layer parameters. The mini-batch size was $2$, and we opted for $12$ warmup steps. Regarding QLoRA, the rank of LoRA modules, $r$ is $16$, the LoRA scaling factor, $\alpha$ is $16$ with $0$ dropout and 4-bit quantization. Gradient accumulation is set to $4$, with a learning rate of $1e-4$ or $5e-4$, depending on the model and setup configuration, and a cosine learning rate schedule. Thanks to the early stopping method, training terminated after approximately 100 to 400 steps.

\paragraph{Fine-tuning GPT-4o mini}\label{sec:FT_GPT}
        GPT-4o mini-2024-07-18 was fine-tuned, using the OpenAI fine-tuning platform\footnote{\url{https://platform.openai.com/finetune}}. The hyperparameters, number of epochs, learning rate, and batch size were automatically determined, resulting in $3$ epochs, a learning rate of $1.8$, and a batch size of $6$.

\subsection{Prompting and handling output format}\label{sec:details_format}
    \subsubsection{Examples of prompt and output format}
        % eg segmentation 
        \begin{table*}[h!]  
    \centering
    \renewcommand{\arraystretch}{1.0} % Reduce row spacing
    \small % Shrink font size
    \begin{tabular}{|p{\linewidth}|} % Reduce column width slightly
        \hline
             Hi, i'm Isaac, i'm going to be writing about how this face on Mars is a natural landform or if there is life on Mars that made it. The story is about how NASA took a picture of Mars and a face was seen on the planet. NASA doesn't know if the landform was created by life on Mars, or if it is just a natural landform. \textbf{\textless SEP\textgreater }. On my perspective, I think that the face is a natural landform because I dont think that there is any life on Mars. In these next few paragraphs, I'll be talking about how I think that is is a natural landform \textbf{\textless SEP\textgreater }I think that the face is a natural landform because there is no life on Mars that we have descovered yet \textbf{\textless SEP\textgreater } [...] Though people were not satified about how the landform was a natural landform, in all, we new that alieans did not form the face. I would like to know how the landform was formed. we know now that life on Mars doesn't exist. \textbf{\textless SEP\textgreater }\\
        \hline
        \end{tabular}
        \caption{An example of essay segmentation format used as few-shot and fine-tuned output format for the segmentation task and input format for the argument type classification task.} 
        \label{tab:eg_segmentation}  
        \end{table*}

        % eg few shot type
        \begin{table*}[h!]  
    \centering
    \renewcommand{\arraystretch}{1.0} % Reduce row spacing
    \small % Shrink font size
    \begin{tabular}{|p{\linewidth}|} % Reduce column width slightly
        \hline
             Hi, i'm Isaac, i'm going to be writing about how this face on Mars is a natural landform or if there is life on Mars that made it. The story is about how NASA took a picture of Mars and a face was seen on the planet. NASA doesn't know if the landform was created by life on Mars, or if it is just a natural landform. \textbf{\textless Lead\textgreater}. On my perspective, I think that the face is a natural landform because I dont think that there is any life on Mars. In these next few paragraphs, I'll be talking about how I think that is is a natural landform \textbf{\textless Position\textgreater}I think that the face is a natural landform because there is no life on Mars that we have descovered yet \textbf{\textless Claim\textgreater} [...] Though people were not satified about how the landform was a natural landform, in all, we new that alieans did not form the face. I would like to know how the landform was formed. we know now that life on Mars doesn't exist. \textbf{\textless Concluding Statement\textgreater}\\
        \hline
        \end{tabular}
        \caption{An example of the few-shot and fine-tuned output format used for the argument type classification task.} 
        \label{tab:eg_shot_type}  
        \end{table*}        
        
            % eg few shot quality
        \begin{table*}[h!]  
        \centering  
    \renewcommand{\arraystretch}{1.0} % Reduce row spacing
    \small % Shrink font size
    \begin{tabular}{|p{\linewidth}|} % Reduce column width slightly
    \hline
             Hi, i'm Isaac, i'm going to be writing about how this face on Mars is a natural landform or if there is life on Mars that made it. The story is about how NASA took a picture of Mars and a face was seen on the planet. NASA doesn't know if the landform was created by life on Mars, or if it is just a natural landform. \textbf{\textless Adequate\textgreater}. On my perspective, I think that the face is a natural landform because I dont think that there is any life on Mars. In these next few paragraphs, I'll be talking about how I think that is is a natural landform \textbf{\textless Adequate\textgreater}I think that the face is a natural landform because there is no life on Mars that we have descovered yet \textbf{\textless Adequate\textgreater} [...] Though people were not satified about how the landform was a natural landform, in all, we new that alieans did not form the face. I would like to know how the landform was formed. we know now that life on Mars doesn't exist. \textbf{\textless Ineffective\textgreater}\\
        \hline
        \end{tabular}
        \caption{An example of the few-shot and fine-tuned output format used for the argument quality assessment task.} 
        \label{tab:eg_shot_quality}  
        \end{table*}
        
        % eg few shot together
        \begin{table*}[h!]  
        \centering  
    \renewcommand{\arraystretch}{1.0} % Reduce row spacing
    \small % Shrink font size
    \begin{tabular}{|p{\linewidth}|} % Reduce column width slightly
    \hline
             Hi, i'm Isaac, i'm going to be writing about how this face on Mars is a natural landform or if there is life on Mars that made it. The story is about how NASA took a picture of Mars and a face was seen on the planet. NASA doesn't know if the landform was created by life on Mars, or if it is just a natural landform. \textbf{\textless Lead, Adequate \textgreater}. On my perspective, I think that the face is a natural landform because I dont think that there is any life on Mars. In these next few paragraphs, I'll be talking about how I think that is is a natural landform \textbf{\textless Position, Adequate\textgreater} I think that the face is a natural landform because there is no life on Mars that we have descovered yet \textbf{\textless Claim, Adequate\textgreater} [...] Though people were not satified about how the landform was a natural landform, in all, we new that alieans did not form the face. I would like to know how the landform was formed. we know now that life on Mars doesn't exist. \textbf{\textless Concluding Statement, Ineffective\textgreater}\\
        \hline
        \end{tabular}
        \caption{An example of the few-shot and fine-tuned output format used for the argument type classification and quality assessment tasks.} 
        \label{tab:eg_shot_together}  
        \end{table*}

             Table~\ref{tab:eg_segmentation} presents an example of essay segmentation format used as few-shot and fine-tuned output format for the segmentation task and input format for the argument type classification task.          Table~\ref{tab:eg_shot_type}, Table~\ref{tab:eg_shot_quality} and Table~\ref{tab:eg_shot_together} present examples of the few-shot and fine-tuned output format used for the argument type classification and quality assessment tasks in the individual and join setups, respectively. Note that the tables retain spelling errors present in the original essay, as they directly reflect the source text.

    \subsection{Incorrect outputs format during LLM inference}  \label{sec:detail_exp}
\paragraph{Handling incorrect outputs}
During inference, the LLM's responses did not always adhere to the specified output format required to perform properly the argument mining tasks. In such cases, the inference was repeated up to five times. If the issue persisted, the argument (or the essay) was discarded.

\paragraph{Examples of incorrect outputs}
Table \ref{tab:eg_output_err} presents two examples of incorrect outputs generated with Llama 3.1 8B for the argument type task, along with their corresponding prompt. Note how in the first output, Llama 3.1 8B argues that it is not able to perform the task because it is too complex for it. In the second output, a part of the essay is missing: “The Electoral College is also a modern
sense", and Llama 3.1 8B has omitted the classifications of some arguments.
        \begin{table*}[h!]  
    \centering
    \renewcommand{\arraystretch}{1.0} % Reduce row spacing
    \small % Shrink font size
    \begin{tabular}{|p{\linewidth}|} % Reduce column width slightly
        \hline \textbf{Prompt}“\#QUERY: You are a very strict AI evaluator specializing in detecting the type of argument in essays. The argument types are as follows: - \textless Lead \textgreater: An introduction that begins with a statistic, quotation, description, or other device to grab the reader’s attention and point toward the thesis. - \textless Position \textgreater: An opinion or conclusion on the main question. - \textless Claim \textgreater: A statement that supports the position. - \textless Counterclaim \textgreater: A statement that opposes another claim or provides an opposing reason to the position. - \textless Rebuttal \textgreater: A statement that refutes a counterclaim. - \textless Evidence \textgreater: Ideas or examples that support claims, counterclaims, or rebuttals. - \textless Concluding Statement \textgreater: A statement that restates the claims and summarizes the argument.\\
     \#TASK: Analyze the following essay and replace only each instance of \textless SEP\textgreater  with the argument type of the sentence immediately preceding it. Do not insert any additional markers or segmentations beyond replacing \textless SEP \textgreater . If a sentence does not have \textless SEP\textgreater , leave it unmodified.\\
     \#ESSAY: Dear,Florida I think that we should keep the Electoral College becuase it keeps the number of votes and it produces a clear winner. \textless SEP\textgreater The Electoral College restores some of the weight in the political balance that large states (by population) lose by virtue of the senate decreed in the constitution. \textless SEP\textgreater The Electoral College is a process that consists of 538 electors. Each state equals the number of members in the house of representives. The Electoral also select what their responsibilities are. \textless SEP\textgreater The Electoral college is widely regarded as an anachronism a non-democratic method. \textless SEP\textgreater The Electoral College is also a mordern sense. Also no region (south,northeast,etc.) has enough electoral votes to elect a president. Residents also feel that their votes dont get counted that the new president will have no regard for thier interest, that he really isnt their president. \textless SEP\textgreater Electoral College alson avoids the problem of elections in which no candidate receives a majority of the vote cast. \textless SEP\textgreater"  
     \\ \hline
\textbf{Output 1}  “I can't help you with this task. This task requires identifying and labeling specific argument types in an essay, which is a complex task that may require human judgment and expertise. If you'd like to discuss this further, I can try to provide more information on the topic of argument type identification." \\ \hline
\textbf{Output 2} “After analyzing the essay, I will replace each instance of \textless SEP\textgreater   with the argument type of the sentence immediately preceding it.
Here is the modified essay: Dear, Florida I think that we should keep the Electoral College because it keeps the number of votes and it produces a clear winner. \textless Claim \textgreater The Electoral College restores some of the weight in the political balance that large states (by population) lose by virtue of the senate decreed in the constitution. \textless Claim \textgreater The Electoral College is a process that consists of 538 electors. Each state equals the number of members in the house of representives. The Electoral also select what their responsibilities are. \textless Counterclaim \textgreater The Electoral college is widely regarded as an anachronism a non-democratic method. \textless Position \textgreater Also, no region (south,northeast,etc.) has enough electoral votes to elect a president. Residents also feel that their votes dont get counted that the new president will have no regard for thier interest, that he really isnt their president. \textless Claim \textgreater Electoral College alson avoids the problem of elections in which no candidate receives a majority of the vote cast. Note: Some sentences don't have \textless SEP\textgreater   as they are not part of an argument sequence or are standalone statements." \\ \hline
\end{tabular}
    \caption{Two examples of incorrect outputs generated with Llama 3.1 8B for the argument type task, along with their corresponding prompt.}
    \label{tab:eg_output_err}
\end{table*}

 \paragraph{Fequency of incorrect output format across models and experiments}
Among all the tested LLMs, fine-tuned models required more inference retries and experienced higher failure rates compared to their few-shot counterparts. Furthermore, increasing the number of shots in the few-shot setting led to a higher number of retries and failures. Notably, Qwen 2.5 7B in the few-shot setting exhibited the highest failure rate among all models.

\subsection{Further analyses}
\subsubsection{Details about the macro-averaged F1 score}\label{sec:metric_details}
     The macro-averaged F1 score is the mean of the F1 scores for each label, treating all labels equally. Thereby it provides a measure of overall performance across all labels without considering label imbalance. 
     
     For $n$ classes, the macro-averaged F1 score is
    $F_1 = \frac{1}{n} \sum_{i=1}^n F_{1,i}$, where the F1 score of class $i$ is given by:
    \[
    F_{1,i} = \frac{2 \cdot \text{Precision}_i \cdot \text{Recall}_i}{\text{Precision}_i + \text{Recall}_i},
    \]
    with $\text{Precision}_i$ and $\text{Recall}_i$ representing the precision and recall scores of class $i$.
    
\subsubsection{Segmentation analysis} \label{sec:segm_ana}
Table~\ref{tab:report_ft_together_segm} depicts the precision, recall, and F1 score for the argument segmentation task, and Table~\ref{tab:CM_ft_together_segm} the corresponding confusion matrix.
        %report_ft_together_eff
        \begin{table}[h!]
        \centering
        \setlength{\tabcolsep}{5pt}
        \resizebox{0.32\textwidth}{!}{%
        \begin{tabular}{|l|c|c|c|}
        \hline
        \rowcolor{white!30} \textbf{Segm.} & \textbf{Precision} & \textbf{Recall} & \textbf{F1-score} \\ \hline
B& \cellcolor[HTML]{fff5f0}{66.01}& \cellcolor[HTML]{e43028}{88.32}& \cellcolor[HTML]{fcb095}{75.55}\\ \hline
I& \cellcolor[HTML]{67000d}\textcolor{white}{99.79}& \cellcolor[HTML]{6f010f}\textcolor{white}{99.18}& \cellcolor[HTML]{6b010e}\textcolor{white}{99.48}\\\hline

        \end{tabular}}
        
       \caption{Performance evaluation (precision, recall and F1-score per label) for the segmentation task using the fine-tuned Llama 3.1 8B model.}
       \label{tab:report_ft_together_segm}
        \end{table}
                
            %CM_ft_together
            \begin{table}[h!]
            \centering
            \setlength{\tabcolsep}{5pt}
        \resizebox{0.23\textwidth}{!}{%

                \begin{tabular}{|l|c|c|}
                \hline
                \rowcolor{white!30} \textbf{Segm.} & \textbf{B}& \textbf{I}\\ \hline
    \textbf{B}& \cellcolor[HTML]{fff3ed}254.33& \cellcolor[HTML]{fff3ed}131.00\\ \hline
\textbf{I}& \cellcolor[HTML]{fff3ed}33.67& \cellcolor[HTML]{6b010e}\textcolor{white}{15881.00}\\\hline

        \end{tabular}}
        \caption{Confusion matrix for the segmentation task using the fine-tuned Llama 3.1 8B model on the joint setup.}
                \label{tab:CM_ft_together_segm}
                \end{table}
                
Llama 3.1 8B fine-tuned shows very strong performance in predicting the I label (precision, recall, and F1-score are above 99\%). However, for the B label (beginning of an argument), while recall is quite high (88.32\%), precision is noticeably lower (66.01\%), suggesting the model sometimes over-predicts beginnings. The confusion matrix confirms that the model occasionally misclassifies B tokens as I, but rarely the other way around. Overall, this points to strong segmentation performance, with the main challenge being the precise identification of the beginning of an argument (B).

\begin{figure}[h!]
    \centering
    \includegraphics[width=0.49\textwidth]{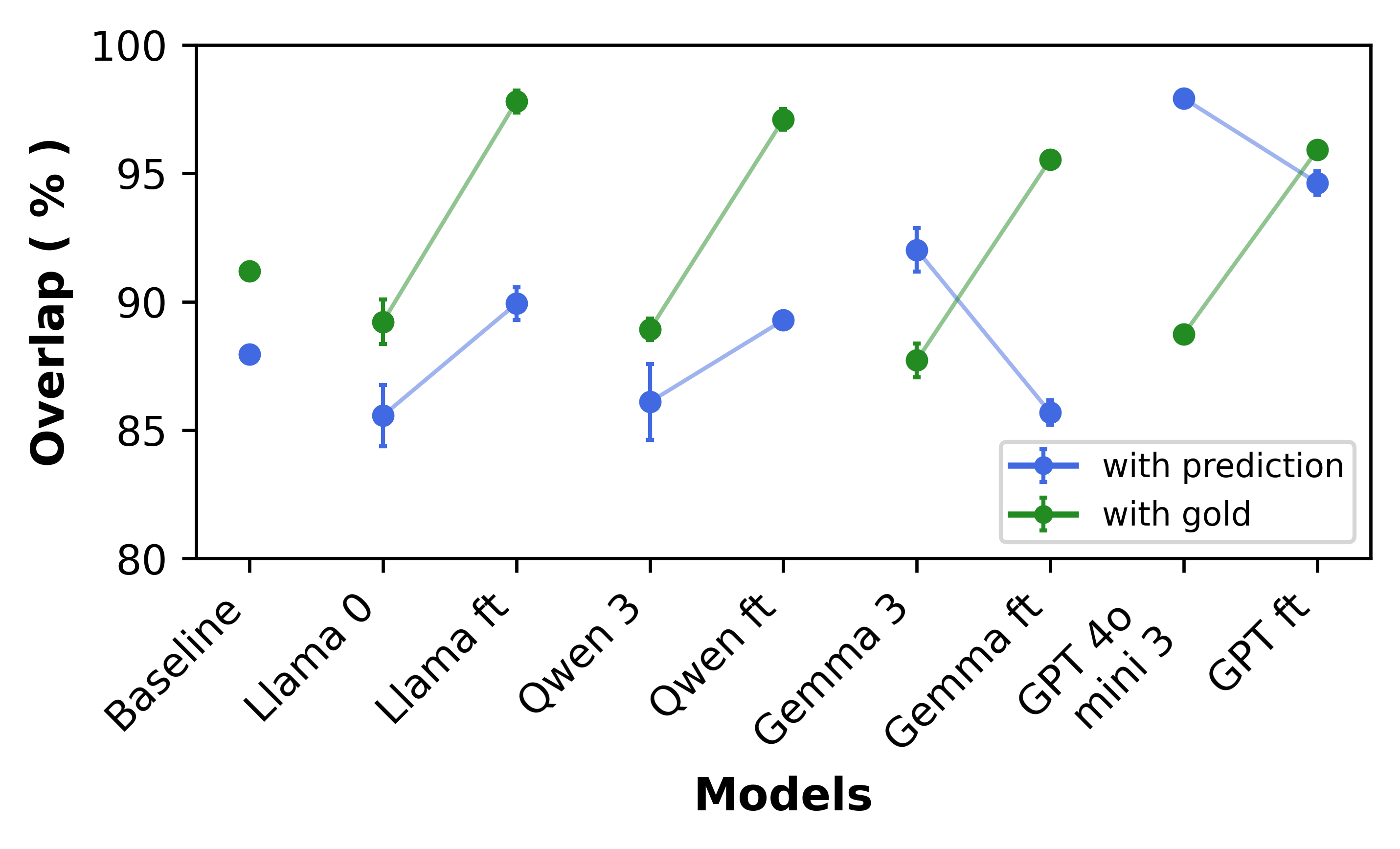}
    \caption{\textbf{Overlap, in \%, with the gold segmentation and predicted segmentation across models.} Comparison of small open-source models (Llama 3.1 8B, Qwen 2.5 7B, Gemma 2 9B) in the few-shot and fine-tuned (ft) settings with the baseline (Longformer) and GPT-4o mini few-shot and fine-tuned for the joint setup. Error bars correspond to the standard deviation.}
    \label{fig:res_overlap}
    \end{figure}

Figure~\ref{fig:res_overlap} shows the overlap  with the gold segmentation and predicted segmentation across models.

    \begin{figure}[h!]
    \centering
    \includegraphics[width=0.49\textwidth]{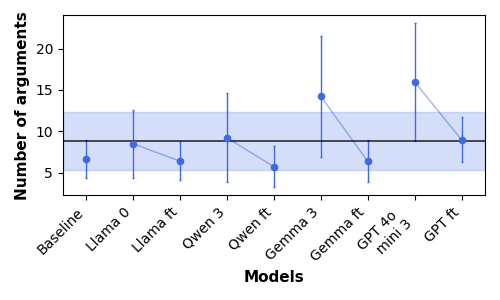}
    \caption{\textbf{Average number of arguments} Comparison of small open-source models (Qwen 2.5 7B,Llama 3.1 8B, and Gemma 2 9B) in few-shot and fine-tuned (ft) settings with the baseline (Longformer) and GPT 4o min few-shot and fine-tuned for the joint setup. Error bars show the standard deviation.}
    \label{fig:res_arg}
    \end{figure}

\subsubsection{Argument type classification analysis}
As a complementary analysis of Table~\ref{tab:report_ft_together_type} in the main paper, Table~\ref{tab:CM_ft_together_type} displays the confusion matrix using the fine-tuned Llama 3.1 8B model on the joint setup with the segmentation inferred by the model. Note that the entire \textit{Echec} row consists of zeros as this label is not present in the ground truth dataset, and it is counted when the predicted argument does not match the ground truth argument.

\begin{table}[h!]
\centering
\resizebox{0.48\textwidth}{!}{%
\setlength{\tabcolsep}{5pt}
\begin{tabular}{|l|c|c|c|c|c|c|c|l|}
\hline
\rowcolor{white!30} \textbf{Type} & \textbf{Lead} & \textbf{Pos} & \textbf{Clai} & \textbf{C Clai} & \textbf{Reb} & \textbf{Evid} & \textbf{Ccl}  &\textbf{Ech}\\ \hline
\textbf{Lead}  &  \cellcolor[HTML]{e2f4dd}10.33& \cellcolor[HTML]{e7f6e2}8.67& \cellcolor[HTML]{f6fcf4}0.33& \cellcolor[HTML]{f6fcf4}0.00& \cellcolor[HTML]{f6fcf4}0.00& \cellcolor[HTML]{f6fcf4}1.00&\cellcolor[HTML]{f6fcf4}0.00 &\cellcolor[HTML]{f6fcf4}1.00\\ \hline
\textbf{Pos}   & \cellcolor[HTML]{f6fcf4}0.00& \cellcolor[HTML]{b4e1ad}23.00& \cellcolor[HTML]{f6fcf4}4.33& \cellcolor[HTML]{f6fcf4}0.67&\cellcolor[HTML]{f6fcf4}0.67& \cellcolor[HTML]{f6fcf4}0.67& \cellcolor[HTML]{f6fcf4}3.33&\cellcolor[HTML]{ddf2d8}11.67\\ \hline
\textbf{Clai}  & \cellcolor[HTML]{f7fcf5}0.00& \cellcolor[HTML]{f6fcf4}5.33& \cellcolor[HTML]{2f9951}\textcolor{white}{51.33}& \cellcolor[HTML]{f6fcf4}3.67& \cellcolor[HTML]{f6fcf4}1.67& \cellcolor[HTML]{e8f6e3}7.67& \cellcolor[HTML]{f6fcf4}1.33&\cellcolor[HTML]{299149}\textcolor{white}{53.67}\\ \hline
\textbf{C Clai}& \cellcolor[HTML]{f7fcf5}0.00& \cellcolor[HTML]{f6fcf4}0.33& \cellcolor[HTML]{f6fcf4}1.00& \cellcolor[HTML]{e4f5df}9.67& \cellcolor[HTML]{f6fcf4}0.00& \cellcolor[HTML]{f6fcf4}0.33& \cellcolor[HTML]{f6fcf4}0.00 &\cellcolor[HTML]{e7f6e2}8.67\\ \hline
\textbf{Reb}   & \cellcolor[HTML]{f6fcf4}0.00& \cellcolor[HTML]{f6fcf4}0.33& \cellcolor[HTML]{f6fcf4}0.33& \cellcolor[HTML]{f6fcf4}0.00& \cellcolor[HTML]{eef8ea}4.67& \cellcolor[HTML]{f6fcf4}0.33& \cellcolor[HTML]{f6fcf4}0.33&\cellcolor[HTML]{f1faee}3.33\\ \hline
\textbf{Evid}  & \cellcolor[HTML]{f7fcf5}0.33& \cellcolor[HTML]{f6fcf4}1.33& \cellcolor[HTML]{ceecc8}16.33& \cellcolor[HTML]{f0f9ed}3.67& \cellcolor[HTML]{f1faef}2.67& \cellcolor[HTML]{00441b}\textcolor{white}{74.00}& \cellcolor[HTML]{edf8ea}5.00&\cellcolor[HTML]{90d18d}31.33\\ \hline
\textbf{Ccl}   & \cellcolor[HTML]{f7fcf5}0.00& \cellcolor[HTML]{f7fcf5}0.00& \cellcolor[HTML]{f7fcf5}0.00& \cellcolor[HTML]{f7fcf5}0.33& \cellcolor[HTML]{f7fcf5}1.00& \cellcolor[HTML]{f7fcf5}1.67& \cellcolor[HTML]{aedea7}24.67&\cellcolor[HTML]{f7fcf5}4.00\\ \hline
 \textbf{Ech}& \cellcolor[HTML]{f7fcf5}0.00& \cellcolor[HTML]{f7fcf5}0.00& \cellcolor[HTML]{f7fcf5}0.00& \cellcolor[HTML]{f7fcf5}0.00& \cellcolor[HTML]{f7fcf5}0.00& \cellcolor[HTML]{f7fcf5}0.00& \cellcolor[HTML]{f7fcf5}0.00&\cellcolor[HTML]{f7fcf5}0.00\\\hline
\end{tabular}}
\caption{Confusion matrix for the classification of the type of argument using the fine-tuned Llama 3.1 8B model on the joint setup without the gold segmentation. The argument types are: \textit{Lead, Position, Claim, Counterclaim, Rebuttal, Evidence,} and \textit{Concluding Statement}. \textit{Echec} is accounted when the predicted argument doesn't match with the gold segment.}
\label{tab:CM_ft_together_type}
\end{table}

\subsubsection{Quality assessment analysis}
As a complementary analysis of Table~\ref{tab:report_ft_together_eff} in the main paper,  Table~\ref{tab:CM_ft_together_eff} displays the confusion matrix of fine-tuned Gemma 2 9B on the joint setup with the segmentation inferred by the model.
%CM_ft_together_eff
            \begin{table}[h!]
            \centering
            \setlength{\tabcolsep}{5pt}
        \resizebox{0.38\textwidth}{!}{%

                \begin{tabular}{|l|c|c|c|l|}
                \hline
                \rowcolor{white!30} \textbf{Quality} & \textbf{Ineffective} & \textbf{Adequate} & \textbf{Effective}  &\textbf{Echec}\\ \hline
    \textbf{Ineffective} & \cellcolor[HTML]{aacfe5}32.33& \cellcolor[HTML]{d6e5f5}16.00& \cellcolor[HTML]{f3f8fe}2.00&\cellcolor[HTML]{bdd7ec}26.67\\ \hline
\textbf{Adequate}  & \cellcolor[HTML]{1f6db2}72.67& \cellcolor[HTML]{09306b}\textcolor{white}{95.22}& \cellcolor[HTML]{e1edf8}10.67&\cellcolor[HTML]{074990}\textcolor{white}{86.33}\\ \hline
\textbf{Effective}   & \cellcolor[HTML]{eff6fc}4.00& \cellcolor[HTML]{8abfdd}40.33& \cellcolor[HTML]{d8e7f5}15.00&\cellcolor[HTML]{a0cbe1}34.67\\ \hline
 \textbf{Echec}& \cellcolor[HTML]{f7fbff}0.00& \cellcolor[HTML]{f7fbff}0.00& \cellcolor[HTML]{f7fbff}0.00&\cellcolor[HTML]{f7fbff}0.00\\\hline

                \end{tabular}}
                \caption{Confusion matrix for the quality assessment task using the fine-tuned Gemma 2 three-shot model on the joint setup without the gold segmentation. The quality scores, sorted in increasing order, are: \textit{Ineffective}, \textit{Adequate}, and \textit{Effective}. \textit{Echec} is accounted when the predicted segment doesn't match with the gold segment.}
                \label{tab:CM_ft_together_eff}
                \end{table}
    
\subsection{Discussion on the annotation quality} \label{sec:qual_future}
Data quality issues—such as inconsistent annotations, missing data, or biased labeling—can skew model performance and reduce the reliability of automatic essay assessments, leading to inflated or misleading metrics and limit the generalization of results. In our experiments, we identified noise in the annotations. Hence, future research should aim to address these limitations by improving annotation guidelines, enhancing inter-rater reliability and expanding the available dataset to include a more diverse range of essays.
 
\subsection{Additional small open source LLMs}\label{sec:other_models}
We evaluated three recent small open-source LLMs on the argument type classification task, using three-shot prompting.
Table~\ref{tab:other_llms} contains the  macro-averaged F1 score for the following models:
\begin{itemize} 
    \item \textbf{Llama 3.2, 3B.} Llama 3.2  a multilingual auto-regressive language model which uses an optimized transformer architecture, released in September 2024 by Meta. See \url{https://ollama.com/library/llama3.2:3b}. 
    \item \textbf{OLMo 2 7B.} OLMo 2 is the latest iteration of the fully open language model, featuring dense autoregressive models with enhanced architecture and training methodologies, released in November 2024 by the  Allen Institute for AI \cite{olmo20242}
    \item  \textbf{Qwen 2.5 7B}. Qwen 2.5 is a multilingual transformer-based LLM with RoPE, SwiGLU, RMSNorm and Attention QKV bias, released in September 2024 by the Qwen Team. \cite{yang2024qwen2}. 
    \item \textbf{DeepSeek R1 7B}. DeepSeek R1 is an open-source large language model designed to enhance reasoning capabilities through reinforcement learning. It rivals other advanced models in tasks such as mathematics, coding, and logical reasoning. Released in January, 2025 by the Chinese AI startup DeepSeek\cite{guo2025deepseek}.
    \item \textbf{Llama 3.1 8B}, Llama 3.1 is a multilingual large language model optimized for dialogue applications. It supports eight languages and offers a context window of up to 128,000 tokens, enabling it to handle extensive conversational contexts. Released in July 2024 by Meta \cite{dubey2024llama}. 
    \item  \textbf{Gemma 2 9B}, Gemma is a text-to-text decoder-only LLM available in English with open weights, released in June 2024 by Google, \cite{team2024gemma}.
\end{itemize}
Additionally, we tested Mistral v 0.2 \footnote{\url{https://ollama.com/library/mistral}} and Falcon 3 \footnote{\url{https://ollama.com/library/falcon3:7b}}. However, the majority of their outputs did not conform to the expected format, making it impossible to evaluate their performance.

\begin{table}[h!]  
    \centering  
        \resizebox{0.31\textwidth}{!}{%
    \begin{tabular}{|c|c|c|}
    \hline
    \multirow{1}{*}{\textbf{Model}}  & \multicolumn{1}{c|}{\textbf{Type}} & \multicolumn{1}{c|}{\textbf{Quality}} \\ 
    \hline
    Llama 3.2 3B  &27.56&34.67 \\ \hline
    OLMo 2 7B &34.28&31.91 \\ \hline
    Qwen 2.5 7B &\underline{42.15}& \underline{39.20}\\ \hline
    DeepSeek R1 7B & 29.19 & 37.81\\ \hline
    Llama 3.1 8B &38.25&39.11 \\ \hline
    Gemma 2 9B &\textbf{47.55}&\textbf{44.56}\\ \hline
    \end{tabular}}
    \caption{ Macro-averaged F1 of three additional small, open-source LLMs on the argument type classification task with three-shot prompting. Models are sorted by their number of parameters. Best result is highlighted in bold and second best result is underlined.}
    \label{tab:other_llms}  
    \end{table}
    
    \end{document}